\documentclass[11pt,a4paper]{article}
\usepackage[hyperref]{emnlp-ijcnlp-2019}
\usepackage{times}
\usepackage{latexsym}
\usepackage{amsmath}
\usepackage{amsfonts}
\usepackage{url}
\usepackage{multirow}
\usepackage{comment}
\usepackage{graphicx}

\aclfinalcopy 

\newcommand{\stitle}[1]{\vspace{0.3ex} \noindent{\bf #1}}

\title{Retrofitting Contextualized Word Embeddings with Paraphrases}

\author{Weijia Shi$^{1}$\thanks{\indent Both authors contributed equally to this work.} , Muhao Chen$^{1,2*}$, Pei Zhou$^{1,3}$ and Kai-Wei Chang$^1$ \\
  $^1$Department of Computer Science, University of California Los Angeles \\
  $^2$Department of Computer and Information Science, University of Pennsylvania \\
  $^3$Department of Computer Science, University of Southern California \\
  {\tt \{swj0419, kwchang\}@cs.ucla.edu}; 
  {\tt muhao@seas.upenn.edu}; 
  {\tt peiz@usc.edu}
} 

\date{}

\begin{document}
\maketitle

\begin{abstract}
Contextualized word embedding models, such as ELMo, generate meaningful representations of words and their context. These models have been shown to have a great impact on downstream applications. However, in many cases, the contextualized embedding of a word changes drastically when the context is paraphrased. 
As a result, the downstream model is not robust to paraphrasing and other linguistic variations.
To enhance the stability of contextualized word embedding models, we propose an approach to retrofitting contextualized embedding models with paraphrase contexts.
  Our method learns an orthogonal transformation on the input space, which seeks to minimize the variance of word representations on paraphrased contexts.
  Experiments show that the retrofitted model significantly outperforms the original ELMo on various sentence classification and language inference tasks.
\end{abstract}
\section{Introduction}
Contextualized word embeddings have shown to be useful for a variety of downstream tasks \cite{peters2018elmo,peters2017semi,mccann2017cove}. 
Unlike traditional word embeddings that represent words with fixed vectors, 
these embedding models encode both words and their contexts and generate context-specific representations. While contextualized embeddings are useful, we observe that a language model-based embedding model, ELMo~\cite{peters2018elmo}, cannot accurately capture the semantic equivalence of contexts. 
Specifically, in cases where the contexts of a word have equivalent or similar meanings but are changed in sentence formation or word order, ELMo may assign very different representations to the word.
Table~\ref{tbl:l2} shows two examples, where ELMo generates very different representations for the boldfaced words under semantic equivalent contexts. 
Quantitatively, 28.3\% of the shared words in the paraphrase sentence pairs on the MRPC corpus \cite{dolan2004msrp} is larger than the average distance between \emph{good} and \emph{bad} in random contexts, and 41.5\% of those exceeds the distance between \emph{large} and \emph{small}.
As a result, the downstream model is not robust to paraphrasing and the performance is hindered.  
{
\begin{table}[t]
\centering
\setlength\tabcolsep{2pt}
\small
\begin{tabular}{|l|c|c|} 
\hline
\textbf{Paraphrased contexts}& \textbf{L2} & \textbf{Cosine}  \\ \hline
\begin{tabular}[c]{@{}l@{}}How can I make \textbf{bigger} my arms? \\ How do I make my arms \textbf{bigger}?\end{tabular}  & 6.42 & 0.27 \\ \hline
\begin{tabular}[c]{@{}l@{}}Some people believe earth is \textbf{flat}. Why?\\ Why do people still believe in \textbf{flat} earth?\end{tabular} & 7.59 &   0.46           \\ \hline
\begin{tabular}[c]{@{}l@{}}It is a very \textbf{small} window. \\ I have a \textbf{large} suitcase. \end{tabular} & 5.44   &    0.26       \\ \hline
\end{tabular}
\caption{
L2 and Cosine distances between embeddings of boldfaced words. The distance between the shared word in the paraphrases is even greater than the distance between
\emph{large} and \emph{small} in random contexts.
}\label{tbl:l2}. 
\vspace{-2em}
\end{table}
}

Infusing the model with the ability to capture the semantic equivalence no doubt benefits semantic-oriented downstream tasks.
Yet, finding an effective solution 
presents key challenges.
First, 
the solution inevitably requires the embedding model to effectively identify paraphrased contexts.
On top of that, the model needs to minimize the difference of a word's representations on paraphrased contexts, without compromising the varying representations on unrelated contexts.
Moreover, the long training time prevents us from redesigning the learning objectives of contextualized embeddings and retraining the model.

To address these challenges,
we propose a simple and effective paraphrase-aware retrofitting (PAR) method 
that is applicable to arbitrary pretrained contextualized embeddings. 
In particular,  PAR prepends an orthogonal transformation layer to a contextualized embedding model. 
Without re-training the parameters of an existing model,
PAR learns the transformation to minimize the difference of the contextualized representations of the shared word in paraphrased contexts,
while differentiating between those in other contexts.
We apply PAR to retrofit ELMo~\cite{peters2018elmo} and show that the resulted embeddings provide more robust contextualized word representations as desired,
which further lead to significant improvements on various sentence classification and inference tasks.

\section{Related Work}
\stitle{Contextualized word embedding models} have been studied by a series of recent research efforts, where different types of pre-trained language models are employed to capture the context information.
CoVe \cite{mccann2017cove} trains a neural machine translation model and extracts representations of input sentences from the source language encoder.
ELMo \cite{peters2018elmo} pre-trains LSTM-based language models from both directions and combines the vectors to construct contextualized word representations.
Recent studies substitute LSTMs 
with Transformers \cite{radford2018gpt1,radford2019language,devlin2019bert}.
As shown in these studies, contextualized word embeddings perform well on downstream tasks at the cost of extensive parameter complexity and the long training process on large corpora \cite{strubell2019energy}.

\stitle{Retrofitting methods} have been used to incorporate semantic knowledge from external resources into word embeddings~\cite{faruqui2014retrofitting, yu2016retrofitting, glavavs2018explicit}. These techniques are shown to improve the characterization of word relatedness and the compositionality of word representations.
To the best of our knowledge, none of the previous approaches has been applied in contextualized word embeddings. 

\section{Paraphrase-Aware Retrofitting}
\begin{figure}[t]
\includegraphics[width=8cm]{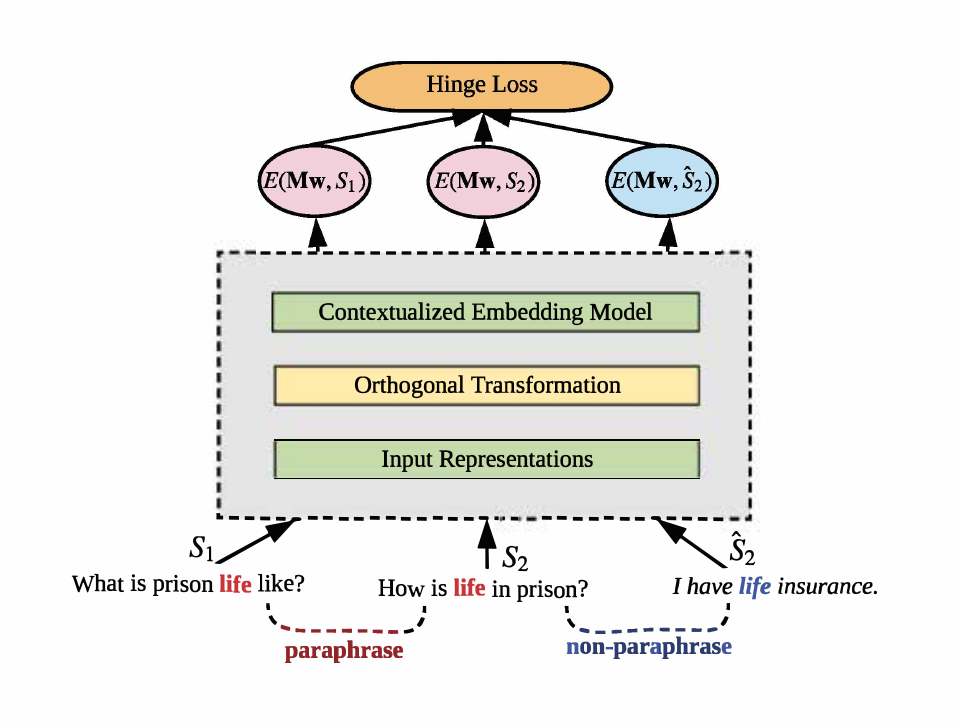}
\centering
\caption{Learning framework of PAR}\label{fig:arch}
\end{figure}

Our method, illustrated in Figure~\ref{fig:arch}, integrates the constraint of the paraphrased context into the contextualized word embeddings by learning the orthogonal transformation on the input space.

\subsection{Contextualized Word Embeddings}
We use $S=(w_1,w_2,\cdots,w_l)$ to denote a sequence of words of length $l$,
where each word $w$ belongs to the vocabulary $V$.
We use boldfaced $\mathbf{w}\in \mathbb{R}^k$ to denote a $k$-dimensional input word embedding,
which can be pre-trained or derived from a character-level encoder (e.g., the character-level CNN used in ELMo \cite{peters2018elmo}).
A contextualized embedding model $E$ takes input vectors of the words in $S$,
and computes the context-specific representation of each word.
The representation of word $w$ specific to the context $S$ is denoted as $E(\mathbf{w}, S)$.


\subsection{Paraphrase-aware Retrofitting}

PAR learns an orthogonal transformation $\mathbf{M} \in \mathbb{R}^{k\times k}$ to reshape the input representation into a specific space,
where the contextualized embedding vectors of a word in paraphrased contexts are collocated, while those in unrelated contexts are differentiated. 
Specifically,
given two contexts $S_1$ and $S_2$ that both contain a shared word $w$,
the contextual difference of a input representation $\mathbf{w}$ is defined by the $L2$ distance,
\begin{equation*}
d_{S_1,S_2}(\mathbf{w})=\left \| E(\mathbf{w}, S_1) - E(\mathbf{w}, S_2) \right \|_2 .
\end{equation*}

\noindent
Let $P$ be the set of paraphrases on the training corpus, we minimize the following hinge loss ($L_H$).
{
\begin{align*}
\sum_{ (S_1, S_2) \in\!  P}\sum_{w\in S_1 \!\cap\! S_2}\!\!\!\left[ d_{S_1,S_2}\!(\mathbf{Mw}\!) \!+\! \gamma \!-\! d_{\hat{S}_1,\hat{S}_2}\!(\mathbf{Mw}\!) \!\right]_+\!\!.
\end{align*}
}

\noindent
$(S_1, S_2) \in  P$ thereof is a pair of paraphrases in $P$. $(\hat{S}_1,\hat{S}_2) \notin P$ is a negative sample generated by 
randomly substituting either $S_1$ or $S_2$ with another sentence in the dataset that contains $w$.
$\gamma > 0$ is a hyper-parameter representing the margin. The operator $[x]_+$ denotes $\max (x, 0)$.

The orthogonalization is realized by the following regularization term.

\begin{equation*}
    L_O=\left \| \mathbf{I} - \mathbf{M}^\top \mathbf{M} \right \|_F,
\end{equation*} 

\noindent
where $\left \| \cdot \right \|_F$ denotes the Frobenius norm, and $\mathbf{I}$ is an identity matrix.
The learning objective of PAR is then denoted as $L=L_H+\lambda L_O$ with a positive hyperparameter $\lambda$.

{
\begin{table*}[t]
\centering
{
\setlength\tabcolsep{1pt}
\small
\begin{tabular}{|l|c|c|c|c|c|c|c|c|c|c|} 
\hline
\multirow{2}{*}{\textbf{Method}}&\multicolumn{4}{c|}{\textbf{Classification (Acc \%) }}&\multicolumn{4}{c|}{\textbf{ Relatedness \textbackslash~Similarity ($\rho$)}}&\multicolumn{2}{c|}{\textbf{Inference (Acc \%)}}\\
\cline{2-11}
&MPQA&MR&CR&SST-2&SICK-R&STS-15&STS-16&STS-Benchmark&MRPC&SICK-E\\
\hline
ELMo (all layers)&89.55&79.72&85.11&86.33&0.84&0.69&0.64&0.65&71.65&81.86\\
ELMo (top layer)&89.30&79.36&84.13&85.28&0.81&0.67&0.63&0.62&70.20&79.64\\
\hline
ELMo-PAR (\emph{MRPC})&92.61&\textbf{83.40}&\textbf{87.01}&86.83&\textbf{0.87}&0.70&0.66&0.64&-&82.89\\
ELMo-PAR (\emph{Sampled Quora})&\textbf{93.76}&81.14&85.52&88.71&0.83&0.71&0.66&0.69&73.22&81.51\\
ELMo-PAR (\emph{PAN})&92.13&83.11&85.73&88.56&0.85&\textbf{0.73}&0.67&\textbf{0.70}&\textbf{74.86}&83.37\\
ELMo-PAR (\emph{PAN+MRPC+Quora})&93.40&82.26&86.39&\textbf{89.26}&0.86&\textbf{0.73}&\textbf{0.68}&0.67&-&\textbf{84.46}\\
\hline 

\end{tabular}
}
\caption{Performance on downstream applications. We report accuracy for classification and inference tasks, and Pearson correlation for relatedness and similarity tasks.
We do not report results of ELMo-PAR on MRPC when MRPC is used in training the model.
The baseline results are from \cite{perone2018evaluation}. }
\label{tbl:result}
\end{table*}
}
{
\begin{table}[t]
\centering
{
\setlength\tabcolsep{1pt}
\small
\begin{tabular}{|l|c|c|c|c|}
\hline
\multirow{2}{*}{\textbf{Model}} & \multicolumn{2}{l|}{\textbf{AddOneSent}} & \multicolumn{2}{l|}{\textbf{AddSent}} \\ \cline{2-5} 
                                & \textbf{EM}       & \textbf{F1}       & \textbf{EM}         & \textbf{F1}        \\ \hline
BiSAE                           & 47.7              & 53.7              & 36.1                & 41.7               \\ \hline
BiSAE-PAR(MRPC)                 & \textbf{51.6}              & \textbf{57.9}              & \textbf{40.8}                & \textbf{47.1}               \\ \hline
\end{tabular}
}
\caption{Exact Match and F1 on Adversarial SQuAD. }\label{tbl:squad}
\end{table}
}

Orthogonalizing $\mathbf{M}$ has two important effects: 
(i) It preserves the word similarity captured by the original input word representation~\cite{rothe2016ultradense};
(ii) It prevents the model from converging to a trivial solution where all word representations collapse to the same embedding vector.

\begin{table*}[t]
\small
\centering
\begin{tabular}{|l|l|l|l|l|l|l|}
\hline
\multirow{2}{*}{\textbf{Model}} & \multicolumn{3}{l|}{\textbf{Paraphrase}}      & \multicolumn{3}{l|}{\textbf{Non-paraphrase}}  \\ \cline{2-7} 
                                & \textbf{MRPC} & \textbf{Quora} & \textbf{PAN} & \textbf{MRPC} & \textbf{Quora} & \textbf{PAN} \\ \hline
ELMo(all layers)                & 3.35          & 3.17           & 4.03         & 3.97          & 4.42           & 6.26         \\ \hline
ELMo-PAR(\textit{PAN+MRPC+Quora})            & 1.80          & 1.34           & 1.21         & 4.73          & 5.49           & 6.54         \\ \hline
\end{tabular}

\caption{Averaging L2 distance for the shared word in paraphrased and non-paraphrased contexts. 
}\label{tbl:dis}

\end{table*}


{
\begin{table}[t]
\centering
\setlength\tabcolsep{2pt}
\small
\begin{tabular}{|l|c|c|} 
\hline
\textbf{Paraphrased contexts}& \textbf{L2} & \textbf{Cosine}  \\ \hline
\begin{tabular}[c]{@{}l@{}}How can I make \textbf{bigger} my arms? \\ How do I make my arms \textbf{bigger}?\end{tabular}  & 2.75 & 0.14 \\ \hline
\begin{tabular}[c]{@{}l@{}}Some people believe earth is \textbf{flat}. Why?\\ Why do people still believe in \textbf{flat} earth?\end{tabular} & 3.29 &   0.16           \\ \hline
\begin{tabular}[c]{@{}l@{}}It is a very \textbf{small} window. \\ I have a \textbf{large} suitcase. \end{tabular} & 5.84   &    0.30       \\ \hline
\end{tabular}
\caption{
L2 and Cosine distance between embeddings of  boldfaced words after retrofitting. 
}\label{tbl:l3}. 
\end{table}
}

\section{Experiment}
Our method can be integrated with any contextualized word embedding models. In our experiment, we apply PAR on ELMo \cite{peters2018elmo} and evaluate the quality of the retrofitted ELMo on a broad range of sentence-level tasks and the adversarial SQuAD corpus.

\subsection{Experimental Configuration}
We use the officially released 3-layer ELMo (original),
which is trained on the 1 Billion Word Benchmark with 93.6 million parameters.
We retrofit ELMo with PAR on the training sets of three paraphrase datasets:
(i) \emph{MRPC} contains 2,753 paraphrase pairs; 
(ii) \emph{Sampled Quora} contains randomly sampled 20,000 paraphrased question pairs~\cite{iyer2017quora}; and
(iii) \emph{PAN} training set~\cite{madnani2012re} contains 5,000 paraphrase pairs.

The orthogonal transformation $\mathbf{M}$ is initialized as an identity matrix.
In our preliminary experiments, we observed that SGD optimizer is more stable and less likely to quickly overfit the training set than other optimizers with adaptive learning rates~\cite{reddi2018convergence,kingma2014adam}.
Therefore, we use SGD with the learning rate of 0.005 and a batch size of 128. 
To determine the terminating condition, we train a Multi-Layer Perceptron (MLP) classifier on the same paraphrase training set and terminate training based on the paraphrase identification performance on a set of held-out paraphrases. The sentence in the dataset is represented by the average of the word embeddings.
$\lambda$ is selected from $\{0.1, 0.5, 1, 2\}$ and $\gamma$ from $\{1, 2, 3, 4\}$ based on validation set. 
The best margin $\gamma$ and epochs $\zeta$ by early stopping are $\{\gamma = 3,$ $\zeta = 20\}$ on MRPC, $\{\gamma = 2,$ $\zeta = 14\}$ on PAN, and \{$\gamma = 3,$ $\zeta = 10\}$ on Sampled Quora, with $\lambda=1$ in all settings. 




\subsection{Evaluation}

We use the SentEval framework~\cite{conneau2018senteval} to evaluate the sentence embeddings on a wide range of sentence-level tasks. We consider two baselines models: (1) ELMo (all layers) constructs a 3,074-dimensional sentence embedding by averaging the hidden states of all the language model layers. (2) ELMo (top layers) encodes a sentence to a 1,024 dimensional vector by averaging the representations of the top layer. We compare these baselines with four variants of PAR built upon ELMo (all layers) that trained on different paraphrase corpora.

\subsection{Task Descriptions}\label{ap:desc}


\textbf{Sentence classification tasks.} 
We evaluate the sentence embedding on four sentence classification tasks including two sentiment analysis (MR~\cite{pang2004sentimental}, SST-2~\cite{socher2013recursive}), product reviews (CR~\cite{hu2004mining}), and opinion polarity (MPQA~\cite{wiebe2005annotating}). 
These tasks are all binary classification tasks. We employ a MLP with a single hidden layer of 50 neurons to train the classifer, using a batch size of 64 and Adam optimizer. 

\noindent
\textbf{Sentence inference tasks} 
We consider two sentence inference tasks: paraphrase identification on MRPC~\cite{dolan2004msrp} and the textual entailment on SICK-E~\cite{marelli2014semeval}. MRPC consists of pairs of sentences, where the model aims to classify if two sentences are semantically equivalent. The SICK dataset contains 10,000 English sentence pairs annotated for relatedness in meaning and entailment. The aim of SICK-E is to detect discourse relations of entailment, contradiction and neutral between the two sentences. Similar to the sentence classification tasks, we apply a MLP with the same hyperparameters to conduct the classification. 

\noindent
\textbf{Semantic textual similarity tasks}. 
Semantic Textual Similarity (STS-15~\cite{agirre2015semeval} and STS-16~\cite{agirre2016semeval}) measures the degree of semantic relatedness of two sentences based on human-labeled scores from 0 to 5. 
We report the Pearson correlation between cosine similarity of two sentence representations and normalized human-label scores. 

\noindent
\textbf{Semantic relatedness tasks} The semantic relatedness tasks include SICK-R~\cite{marelli2014semeval} and the STS Benchmark dataset~\cite{cer2017semeval}, 
which comprise pairs of sentences annotated with semantic scores between 0 and 5. The goal of the tasks is to measure the degree of semantic relatedness between two sentences. We learn tree-structured LSTM~\cite{tai2015improved} to predict the probability distribution of relatedness scores. 

\noindent
\textbf{Adversarial SQuAD} 
The Stanford Question Answering Datasets (SQuAD)~\cite{rajpurkar2016squad} is a machine comprehension dataset containing 107,785 human-generated reading comprehension questions annotated on Wikipedia articles. Adversarial SQuAD~\cite{jia2017adversarial} appends adversarial sentences to the passage in the SQuAD dataset to study the robustness of the model.
We conduct evaluations on two Adversarial SQuAD datasets: AddOneSent which adds a random human-approved sentence, and AddSent which adds grammatical sentences that look similar to the question. We train the Bi-Directional Attention Flow (BiDAF) network~\cite{seo2016bidirectional} with self-attention and ELMo embeddings on the SQuAD dataset and test it on the adversarial SQuAD datasets.

\subsection{Result Analysis}
The results reported in Table~\ref{tbl:result} show that PAR leads to 2\% $\sim$ 4\% improvement in accuracy on sentence classification tasks and sentence inference tasks.
It leads to 0.03 $\sim$ 0.04 improvement in Pearson correlation ($\rho$) on semantic relatedness and textual similarity tasks.
The improvements on sentence similarity and semantic relatedness tasks shows that ELMo-PAR is more stable to the semantic-preserving modifications but more sensitive to subtle yet semantic-changing perturbations. 
PAR model trained on the combined corpus (\emph{PAN+MRPC+Sampled Quora}) achieves the best improvement across all these tasks, showing the model benefits from a larger paraphrase corpus. Besides sentence-level tasks, Table~\ref{tbl:squad} shows that the proposed PAR method notably improves the performance of a downstream question-answering task. 
For AddSent, ELMo-PAR achieves 40.8\% in EM and 47.1\% in F1. For AddOneSent, it boosts EM to 51.6\% and F1 to 57.9\%, which clearly shows that the proposed PAR method enhances the robustness of the downstream model combined with ELMo.

\subsection{Case Study}

\stitle{Shared word distances} 
We compute the average embedding distance of shared words in paraphrase and non-paraphrase sentence pairs from test sets of MRPC, PAN, and Quora. Results are listed in Table~\ref{tbl:dis}. Table~\ref{tbl:l3} shows the ELMo-PAR embedding distance for the shared words in the examples in Table~\ref{tbl:l2}.
Our model effectively minimizes the embedding distance of the shared words in the paraphrased contexts and maximize such distance in the non-paraphrased contexts.

\section{Conclusion}
We propose a method for retrofitting contextualized word embeddings, which leverages semantic equivalence information from paraphrases.
PAR learns an orthogonal transformation on the input space of an existing model by minimizing the contextualized representations of shared words on paraphrased contexts without compromising the varying representations on non-paraphrased contexts. We demonstrate the effectiveness of this method applied to ELMo by a wide selection of semantic tasks.
We seek to extend the use of PAR to other contextualized embeddings \cite{devlin2019bert,mccann2017cove} in future work.

\section{Acknowledgement}

This work was supported in part by National Science Foundation Grant IIS-1760523. We thank reviewers for their comments.

\bibliographystyle{acl_natbib}
\bibliography{main}

\begin{thebibliography}{30}
\expandafter\ifx\csname natexlab\endcsname\relax\def\natexlab#1{#1}\fi

\bibitem[{Agirre et~al.(2015)Agirre, Banea, Cardie, Cer, Diab, Gonzalez-Agirre,
  Guo, Lopez-Gazpio, Maritxalar, Mihalcea et~al.}]{agirre2015semeval}
Eneko Agirre, Carmen Banea, Claire Cardie, Daniel Cer, Mona Diab, Aitor
  Gonzalez-Agirre, Weiwei Guo, Inigo Lopez-Gazpio, Montse Maritxalar, Rada
  Mihalcea, et~al. 2015.
\newblock Semeval-2015 task 2: Semantic textual similarity, english, spanish
  and pilot on interpretability.
\newblock In \emph{Proceedings of the 9th international workshop on semantic
  evaluation}.

\bibitem[{Agirre et~al.(2016)Agirre, Banea, Cer, Diab, Gonzalez-Agirre,
  Mihalcea, Rigau, and Wiebe}]{agirre2016semeval}
Eneko Agirre, Carmen Banea, Daniel Cer, Mona Diab, Aitor Gonzalez-Agirre, Rada
  Mihalcea, German Rigau, and Janyce Wiebe. 2016.
\newblock Semeval-2016 task 1: Semantic textual similarity, monolingual and
  cross-lingual evaluation.
\newblock In \emph{Proceedings of the 10th International Workshop on Semantic
  Evaluation}.

\bibitem[{Cer et~al.(2017)Cer, Diab, Agirre, Lopez-Gazpio, and
  Specia}]{cer2017semeval}
Daniel Cer, Mona Diab, Eneko Agirre, Inigo Lopez-Gazpio, and Lucia Specia.
  2017.
\newblock Semeval-2017 task 1: Semantic textual similarity multilingual and
  crosslingual focused evaluation.
\newblock In \emph{Proceedings of the 11th International Workshop on Semantic
  Evaluation}.

\bibitem[{Conneau and Kiela(2018)}]{conneau2018senteval}
Alexis Conneau and Douwe Kiela. 2018.
\newblock Senteval: An evaluation toolkit for universal sentence
  representations.
\newblock In \emph{Proceedings of the Eleventh International Conference on
  Language Resources and Evaluation (LREC-2018)}.

\bibitem[{Devlin et~al.(2019)Devlin, Chang, Lee, and
  Toutanova}]{devlin2019bert}
Jacob Devlin, Ming-Wei Chang, Kenton Lee, and Kristina Toutanova. 2019.
\newblock Bert: Pre-training of deep bidirectional transformers for language
  understanding.
\newblock In \emph{NAACL}.

\bibitem[{Dolan et~al.(2004)Dolan, Quirk, and Brockett}]{dolan2004msrp}
Bill Dolan, Chris Quirk, and Chris Brockett. 2004.
\newblock Unsupervised construction of large paraphrase corpora: Exploiting
  massively parallel news sources.
\newblock In \emph{COLING}.

\bibitem[{Faruqui et~al.(2015)Faruqui, Dodge, Jauhar, Dyer, Hovy, and
  Smith}]{faruqui2014retrofitting}
Manaal Faruqui, Jesse Dodge, Sujay~K Jauhar, Chris Dyer, Eduard Hovy, and
  Noah~A Smith. 2015.
\newblock Retrofitting word vectors to semantic lexicons.
\newblock In \emph{NAACL}.

\bibitem[{Glava{\v{s}} and Vuli{\'c}(2018)}]{glavavs2018explicit}
Goran Glava{\v{s}} and Ivan Vuli{\'c}. 2018.
\newblock Explicit retrofitting of distributional word vectors.
\newblock In \emph{ACL}.

\bibitem[{Hu and Liu(2004)}]{hu2004mining}
Minqing Hu and Bing Liu. 2004.
\newblock Mining and summarizing customer reviews.
\newblock In \emph{Proceedings of the tenth ACM SIGKDD international conference
  on Knowledge discovery and data mining}.

\bibitem[{Iyer et~al.(2017)Iyer, Dandekar, and Csernai}]{iyer2017quora}
Shankar Iyer, Nikhil Dandekar, and Korn{\'e}l Csernai. 2017.
\newblock First quora dataset release: Question pairs.
\newblock \emph{data.quora.com}.

\bibitem[{Jia and Liang(2017)}]{jia2017adversarial}
Robin Jia and Percy Liang. 2017.
\newblock Adversarial examples for evaluating reading comprehension systems.
\newblock In \emph{EMNLP}.

\bibitem[{Kingma and Ba(2015)}]{kingma2014adam}
Diederik~P Kingma and Jimmy Ba. 2015.
\newblock Adam: A method for stochastic optimization.
\newblock \emph{ICLR}.

\bibitem[{Madnani et~al.(2012)Madnani, Tetreault, and Chodorow}]{madnani2012re}
Nitin Madnani, Joel Tetreault, and Martin Chodorow. 2012.
\newblock Re-examining machine translation metrics for paraphrase
  identification.
\newblock In \emph{NAACL}.

\bibitem[{Marelli et~al.(2014)Marelli, Bentivogli, Baroni, Bernardi, Menini,
  and Zamparelli}]{marelli2014semeval}
Marco Marelli, Luisa Bentivogli, Marco Baroni, Raffaella Bernardi, Stefano
  Menini, and Roberto Zamparelli. 2014.
\newblock Semeval-2014 task 1: Evaluation of compositional distributional
  semantic models on full sentences through semantic relatedness and textual
  entailment.
\newblock In \emph{Proceedings of the 8th international workshop on semantic
  evaluation}.

\bibitem[{McCann et~al.(2017)McCann, Bradbury, Xiong, and
  Socher}]{mccann2017cove}
Bryan McCann, James Bradbury, Caiming Xiong, and Richard Socher. 2017.
\newblock Learned in translation: Contextualized word vectors.
\newblock In \emph{NIPS}.

\bibitem[{Pang and Lee(2004)}]{pang2004sentimental}
Bo~Pang and Lillian Lee. 2004.
\newblock A sentimental education: Sentiment analysis using subjectivity
  summarization based on minimum cuts.
\newblock In \emph{ACL}.

\bibitem[{Perone et~al.(2018)Perone, Silveira, and
  Paula}]{perone2018evaluation}
Christian~S Perone, Roberto Silveira, and Thomas~S Paula. 2018.
\newblock Evaluation of sentence embeddings in downstream and linguistic
  probing tasks.
\newblock \emph{arXiv preprint arXiv:1806.06259}.

\bibitem[{Peters et~al.(2017)Peters, Ammar, Bhagavatula, and
  Power}]{peters2017semi}
Matthew Peters, Waleed Ammar, Chandra Bhagavatula, and Russell Power. 2017.
\newblock Semi-supervised sequence tagging with bidirectional language models.
\newblock In \emph{ACL}.

\bibitem[{Peters et~al.(2018)Peters, Neumann, Iyyer, Gardner, Clark, Lee, and
  Zettlemoyer}]{peters2018elmo}
Matthew Peters, Mark Neumann, Mohit Iyyer, Matt Gardner, Christopher Clark,
  Kenton Lee, and Luke Zettlemoyer. 2018.
\newblock Deep contextualized word representations.
\newblock In \emph{NAACL}.

\bibitem[{Radford et~al.(2018)Radford, Narasimhan, Salimans, and
  Sutskever}]{radford2018gpt1}
Alec Radford, Karthik Narasimhan, Tim Salimans, and Ilya Sutskever. 2018.
\newblock Improving language understanding by generative pre-training.

\bibitem[{Radford et~al.(2019)Radford, Wu, Child, Luan, Amodei, and
  Sutskever}]{radford2019language}
Alec Radford, Jeffrey Wu, Rewon Child, David Luan, Dario Amodei, and Ilya
  Sutskever. 2019.
\newblock Language models are unsupervised multitask learners.

\bibitem[{Rajpurkar et~al.(2016)Rajpurkar, Zhang, Lopyrev, and
  Liang}]{rajpurkar2016squad}
Pranav Rajpurkar, Jian Zhang, Konstantin Lopyrev, and Percy Liang. 2016.
\newblock Squad: 100,000+ questions for machine comprehension of text.
\newblock In \emph{EMNLP}.

\bibitem[{Reddi et~al.(2018)Reddi, Kale, and Kumar}]{reddi2018convergence}
Sashank~J Reddi, Satyen Kale, and Sanjiv Kumar. 2018.
\newblock On the convergence of adam and beyond.
\newblock In \emph{ICLR}.

\bibitem[{Rothe et~al.(2016)Rothe, Ebert, and
  Sch{\"u}tze}]{rothe2016ultradense}
Sascha Rothe, Sebastian Ebert, and Hinrich Sch{\"u}tze. 2016.
\newblock Ultradense word embeddings by orthogonal transformation.
\newblock In \emph{NAACL}.

\bibitem[{Seo et~al.(2017)Seo, Kembhavi, Farhadi, and
  Hajishirzi}]{seo2016bidirectional}
Minjoon Seo, Aniruddha Kembhavi, Ali Farhadi, and Hannaneh Hajishirzi. 2017.
\newblock Bidirectional attention flow for machine comprehension.
\newblock In \emph{ICLR}.

\bibitem[{Socher et~al.(2013)Socher, Perelygin, Wu, Chuang, Manning, Ng, and
  Potts}]{socher2013recursive}
Richard Socher, Alex Perelygin, Jean Wu, Jason Chuang, Christopher~D Manning,
  Andrew Ng, and Christopher Potts. 2013.
\newblock Recursive deep models for semantic compositionality over a sentiment
  treebank.
\newblock In \emph{EMNLP}.

\bibitem[{Strubell et~al.(2019)Strubell, Ganesh, and
  McCallum}]{strubell2019energy}
Emma Strubell, Ananya Ganesh, and Andrew McCallum. 2019.
\newblock Energy and policy considerations for deep learning in nlp.
\newblock In \emph{ACM}.

\bibitem[{Tai et~al.(2015)Tai, Socher, and Manning}]{tai2015improved}
Kai~Sheng Tai, Richard Socher, and Christopher~D Manning. 2015.
\newblock Improved semantic representations from tree-structured long
  short-term memory networks.
\newblock In \emph{ACL-IJCNLP}.

\bibitem[{Wiebe et~al.(2005)Wiebe, Wilson, and Cardie}]{wiebe2005annotating}
Janyce Wiebe, Theresa Wilson, and Claire Cardie. 2005.
\newblock Annotating expressions of opinions and emotions in language.
\newblock \emph{Language resources and evaluation}.

\bibitem[{Yu et~al.(2016)Yu, Cohen, Wallace, Bernstam, and
  Johnson}]{yu2016retrofitting}
Zhiguo Yu, Trevor Cohen, Byron Wallace, Elmer Bernstam, and Todd Johnson. 2016.
\newblock Retrofitting word vectors of mesh terms to improve semantic
  similarity measures.
\newblock In \emph{Proceedings of the Seventh International Workshop on Health
  Text Mining and Information Analysis}.

\end{thebibliography}

\end{document}